\documentclass[10pt,letterpaper]{article}

\usepackage{cogsci}
\usepackage{amsmath}
\usepackage{amsfonts}
\usepackage{graphicx}
\usepackage{makecell}
\usepackage[style=apa,sortcites=true,sorting=nyt,backend=biber, uniquename=false]{biblatex}
\cogscifinalcopy 

\usepackage{pslatex}
\usepackage{float} 

\title{Incoherent Probability Judgments in Large Language Models}
 
\author{{\large \bf Jian-Qiao Zhu (jz5204@princeton.edu)} \\
  Department of Computer Science \\
  Princeton University
  \And {\large \bf Thomas L. Griffiths (tomg@princeton.edu)} \\
  Department of Psychology and Computer Science \\
  Princeton University}

\begin{document}

\maketitle

\begin{abstract}
Autoregressive Large Language Models (LLMs) trained for next-word prediction have demonstrated remarkable proficiency at producing coherent text. But are they equally adept at forming coherent probability judgments? We use probabilistic identities and repeated judgments to assess the coherence of probability judgments made by LLMs. Our results show that the judgments produced by these models are often incoherent, displaying human-like systematic deviations from the rules of probability theory. Moreover, when prompted to judge the same event, the mean-variance relationship of probability judgments produced by LLMs shows an inverted-U-shaped like that seen in humans. We propose that these deviations from rationality can be explained by linking autoregressive LLMs to implicit Bayesian inference and drawing parallels with the Bayesian Sampler model of human probability judgments. 

\end{abstract}

\section{Introduction}

While Large Language Models (LLMs) are primarily trained to predict the next word in text, they have demonstrated a remarkable breadth in their capabilities across various cognitive tasks and domains \parencite{bubeck2023sparks}. The success of LLMs presents new opportunities to deepen our understanding of both machine and human intelligence. Notably, LLMs can generate long coherent passages of texts, although they occasionally produce fabricated information, a phenomenon referred to as ``hallucination'' \parencite{radford2019language}. While hallucination predominantly affects the factual accuracy of the output, it generally does not compromise its coherence. This raises an interesting question: are the judgments made by LLMs also coherent, even when they are factually incorrect?

\begin{figure*}[t!]
\centering
\includegraphics[width=0.98\textwidth]{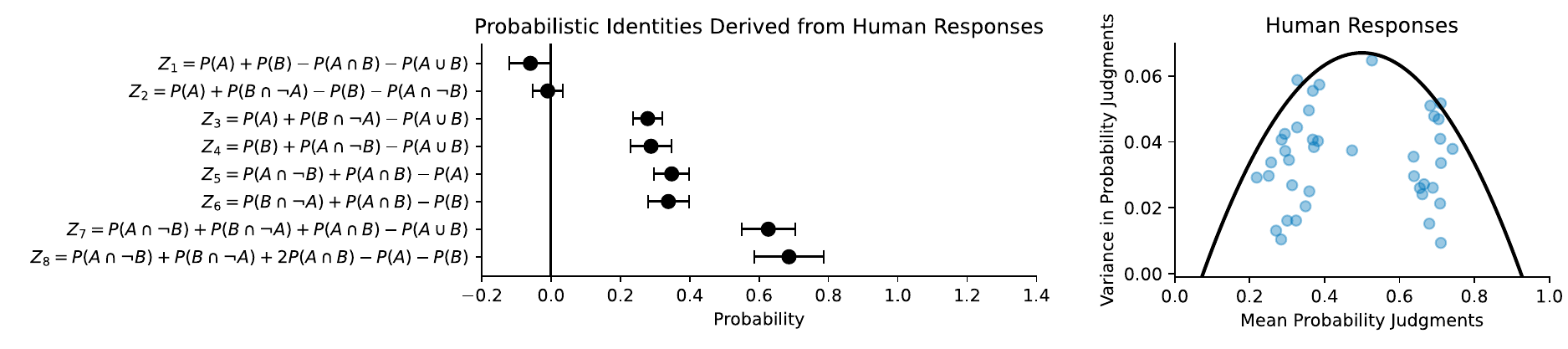}
\caption{Bias and variability in human probability judgments as revealed by \textbf{(left)} probabilistic identities and \textbf{(right)} mean-variance relationship. Error bars are 95\% CI. Solid line represents the best-fitting linear regression. Data adapted from \textcite{zhu2020bayesian} Experiment 1. }
\label{fig:human_prob}
\end{figure*}

In this paper, we focus on evaluating probability judgments -- a domain where coherence can be rigorously assessed against the rules of probability theory. We tasked LLMs with estimating the probability of various future events set in the year 2025 (see Table \ref{tab:events} for details). To quantitatively assess the incoherence of these probability judgments, we employed a series of probabilistic identities, comprehensively listed in Figure \ref{fig:human_prob}. According to these identities, coherent judgments should invariably yield a value of zero. Any deviation from this benchmark serves as a measure of incoherence in the LLMs' probability judgments. This approach aligns with the framework proposed by \textcite{costello2014surprisingly}, who have previously used these probabilistic identities to evaluate human probabilistic reasoning. Moreover, as LLMs generate a probability distribution over probability judgments, we can also examine the stochasticity in these judgments (i.e., consistency over time/information) by repeatedly resubmitting the exact same prompt to evaluate the variability in responses.

To preview our results, we found that four state-of-the-art LLMs varying in training dataset size, computational resources, and the number of model parameters consistently exhibit biases in probabilistic identities that mirror those found in human cognition \parencite{costello2014surprisingly, zhu2020bayesian, huang2023quantum}. Furthermore, when queries about the same event are made repeatedly at temperature setting 1.0 (i.e., when the outputs of LLMs are based on the model's learned probabilities), we find that the mean and the variance of probability judgments calculated over repetitions display a pattern resembling an inverted-U shape, akin to the mean-variance relationships derived from human probability judgments \parencite{sundh2023unified, zhu2023autocorrelated}.

We propose that the human-like patterns in probability judgments produced by the LLMs may stem from the use of the autoregressive training objectives (cf.~\cite{mccoy2023embers}). To understand this phenomenon, we examined two prominent computational models that explain human probability judgments: the Probability Theory plus Noise (PT+N) model \parencite{costello2014surprisingly} and the Bayesian Sampler model \parencite{zhu2020bayesian}. Although the LLMs' responses do not qualitatively differentiate between these two models, we argue that the responses align more closely with the Bayesian Sampler model. This model, which suggests responses are guided by Bayesian inference, appears to better capture the neural networks' behavior than the PT+N model's characterization based on the probability theory distorted by noise. This hypothesis is bolstered by drawing parallels between the autoregressive processes in LLMs and implicit Bayesian inference mechanisms, as recently argued for in machine learning \parencite{xie2021explanation, zhang2023deep, griffiths2023bayes}.

\begin{table*}[t!]
\setlength{\tabcolsep}{2pt}
    \centering
    \caption{Twelve weather-related and twelve political pairs of events posed to the LLMs. }
    \label{tab:events}
    \begin{tabular}{cccc}
       Weather A & Weather B  & Politics A & Politics B\\ \hline
         Rainy & Cold & Britain left the EU & Greece left the EU\\
         Cloudy & Stormy & American cars increase & Petrol price increases \\
         Chilly & Thundery & Climate change impacts American weather & World greenhouse gas emissions are reduced\\
         Hot & Windy & The US is at war in the Middle East & Major terrorist attack occurs in the US\\
         Sunny & Misty & Europe grows poorer & Unemployment in Europe rises above 20\%\\
         Foggy & Wet & Hurricanes and typhoons are more frequent & Average world temperature increases \\
         Freezing & Humid & Generative AI is a trillion-dollar market & AI-designed antibiotics are available on prescription \\
         Drizzly & Breezy & All television become Internet-based & AI has made full-length movies\\
         Hazy & Warm & Tech unemployment has risen & The divide in income levels has expanded \\
         Icy & Snowy & Cities ban fossil-fuel vehicles & One-third of new cars are electric \\
         Breezy & Dry & Depression becomes the No.1 disease burden & Manufacturing jobs disappear in the West\\
         Normal & Typical & The majority of UK homes are rented & Married couples are a minority in the UK\\
       \hline
    \end{tabular}
\end{table*}

\section{Background}
There is a rich history of exploring biases in human probabilistic reasoning, a line of inquiry that, arguably, dates back to the early 19th century with the work of Laplace \parencite{miller2020laplace} and become influential via the ``heuristics and biases'' research program \parencite{kahneman1972subjective, gigerenzer2011heuristic}. Here, rather than testing specific biases or fallacies in LLMs, as has been explored in \textcite{binz2023using} and \textcite{horton2023large}, we choose to focus on analyzing the statistical properties of LLMs' probability judgments for a range of events. This can be achieved through (i) the combination of individual probability judgments into probabilistic identities, and (ii) the repeated elicitation of probability judgments for identical events.

\subsection{Probabilities in Large Language Models}
Autoregressive LLMs are trained to predict the next word, where the training distribution is typically based on a large sample of Internet text. These models typically use a neural network architecture known as the Transformer \parencite{vaswani2017attention}. To predict the next word, LLMs compute a conditional probability given the preceding sequence of $n$ words: $p(w_{n+1}|w_{1:n})$. The precision of this distribution is controlled by the temperature parameter. When the temperature is set low (i.e., close to 0), the model is more likely to pick the most probable next word based on its training. A higher temperature setting makes the model's word choices more equal in probability, leading to more diverse outputs.

As AI systems based on LLMs are becoming more capable of interacting with people in real-world applications, AI researchers have been studying the probabilities that these systems predict for different events (e.g., \cite{ liang2022holistic, jiang2012calibrating}). The primary objective within AI and machine learning is to calibrate these AI-generated probability estimates with true frequencies \parencite{kumar2019verified, shrivastava2023llamas, zadrozny2001obtaining}. While it is vital to ensure that the probability judgments produced by AI are well-calibrated, the coherence of probability judgments is equally important. In fact, there are strong theoretical reasons linking coherence and accuracy in boundedly rational agents \parencite{zhu2022clarifying}.

It is true that accurate probability judgments (i.e., those that perfectly reflect true frequencies) are also coherent. However, the reverse is not always true: coherent probability judgments are not necessarily more accurate. It has been proved that, for any set of incoherent probability judgments, there exists at least one set of coherent probability judgments that better correspond to the true frequencies \parencite{leitgeb2010objective}.

\subsection{Assessing Coherence via Probabilistic Identities}
Coherent probability judgments should be consistent when applied to events that are logically related. This type of coherence can be studied using probabilistic identities, which are formulated by combining individual probability judgments from a pair of binary events, $A$ and $B$ \parencite{costello2014surprisingly, zhu2020bayesian}. In typical experimental setups, participants are asked about either single events (e.g., $P(A)$), conjunctions of the events (e.g., $P(A \cap B)$), or disjunctions of the events (e.g., $P(A \cup B)$). These individual probability judgments are asked independently and subsequently combined to construct probabilistic identities. For example, the identity $Z_1$ is composed of four distinct probability judgments: $Z_1 = P(A)+P(B)-P(A\cap B)-P(A\cup B)$. A key feature of these identities is that they should all equal to zero, according to probability theory.

These identities serve as instrumental tools for quantifying the degree of incoherence in an individual's probability judgments. An individual strictly following the rules of probability theory should make all such identities zero. Moreover, when an agent uses an unbiased estimate of the underlying probabilities (e.g., by drawing i.i.d. samples and using relative frequency for probability estimation), the identities derived from her responses should also equal zero, on average.

When these probabilistic identities are tested on human participants (see Figure \ref{fig:human_prob}, left panel), the results consistently show that the extent of the deviations from zero appears to be influenced by the balance of positive and negative terms within each probabilistic identity \parencite{costello2014surprisingly, zhu2020bayesian, chater2020probabilistic}. For instance, identities $Z_1$ and $Z_2$, which each contain an equal number of positive and negative terms (two of each), typically yield mean responses that approximate zero. In contrast, identities $Z_3$ through $Z_6$, which have one more positive term than negative, demonstrate a different pattern in mean responses. This disparity becomes more pronounced in identities $Z_7$ and $Z_8$, where there are two more positive terms than negative. Consequently, the mean responses for  $Z_7$ and $Z_8$ are observed to be twice as large as those for $Z_3$ through $Z_6$.

It is important to note that if an individual's probability judgments remain constant across various queries (for instance, consistently responding with a probability of 0.5 for all queries), their probabilistic identities would deviate in a manner similar to that observed in human judgments. However, it is rarely the case that individuals respond to all queries with a fixed probability value. In reality, even when presented with the same query on different occasions, individuals often exhibit variability in their probability judgments \parencite{sundh2023unified}. This is also true of LLMs, which generate stochastic responses even when presented with identical prompts.

\subsection{Repeated Judgments}

Coherent probability judgments should also remain consistent as long as the underlying information or evidence hasn't changed. If one judges the probability of an event at one moment, a coherent reassessment at a later time without new information should yield the same probability. This kind of self-consistency is studied using repeated judgments, as variability in probability judgments is important for understanding the underlying mechanisms that people use to formulate these judgments.

In economics and psychology, there have been numerous experiments where individuals are asked to repeat the same task at different times, sometimes even when these occasions are close together in time (e.g., \cite{mosteller1951experimental, zhu2020bayesian, sundh2023unified}). This has led to the observation of significant fluctuations in behavior. For instance, preferences between different monetary gambles have been noted to display marked variability within short time frames \parencite{mosteller1951experimental, loomes1998testing}. Moreover, people's probability judgments of the same event have been found to differ from one instance to another, even in the absence of any new information influencing these judgments  \parencite{sundh2023unified, zhu2020bayesian}.

In addition to this observed variability, certain structures have been identified using repeated probability judgments. Notably, it has been observed that probability judgments closer to 0.5 exhibit greater variability compared to those near the extremities of 0 or 1 \parencite{sundh2023unified}. This phenomenon renders an inverted-U-shaped relationship between the mean and variance of repeated judgments (see Figure \ref{fig:human_prob}, right panel). This inverted-U-shaped curve also demonstrates variation across different participants and events \parencite{sundh2023unified}. Generally, a downward shift in the curve indicates a reduction in the variability of repeated judgments; an inward shift of the curve suggests a `shrinkage effect,' which has been theorized to occur when employing a `stronger' prior to moderate all judgments \parencite{zhu2023autocorrelated}.

\begin{figure*}[t!]
\centering
\includegraphics[width=\textwidth]{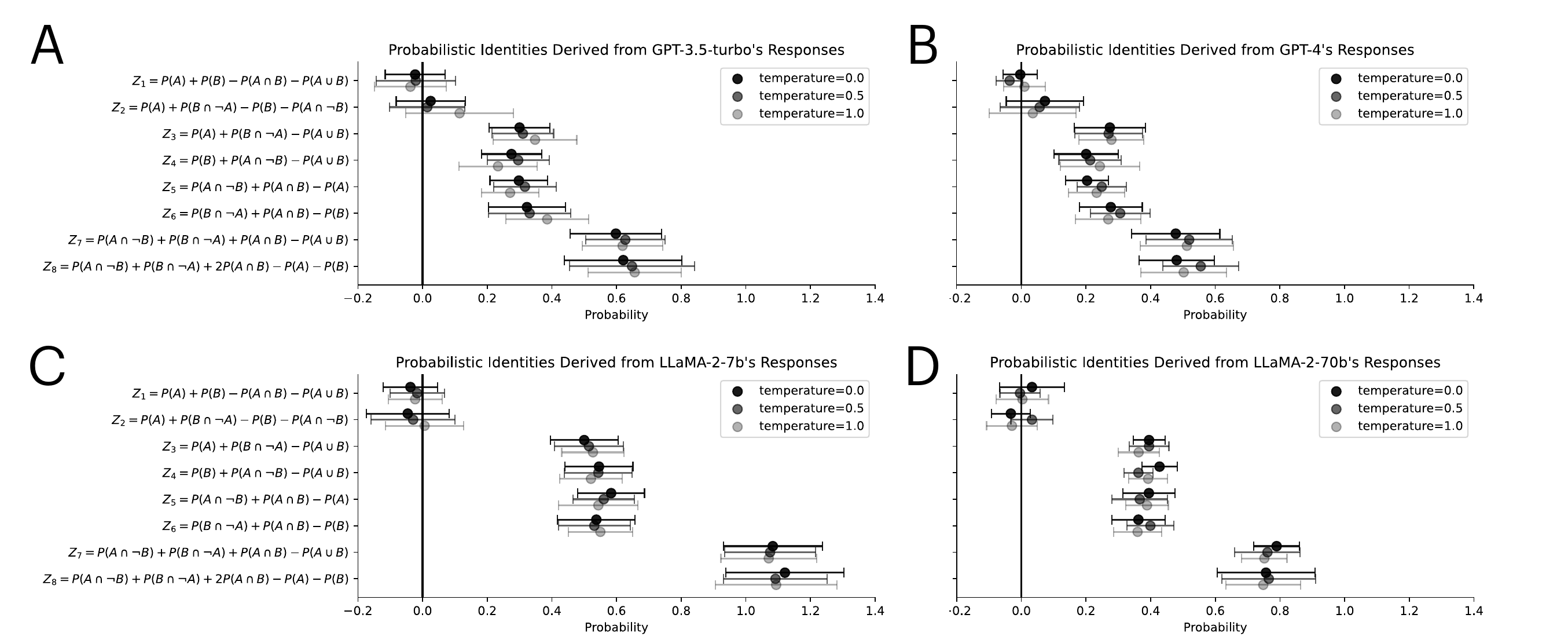}
\caption{Probabilistic identities based on LLM responses. For coherent judgments, all probabilistic identities should be zero. \textbf{(A)} GPT-3.5-turbo model. \textbf{(B)} GPT-4 model. \textbf{(C)} LLaMA-2 model with 7b parameters. \textbf{(D)} LLaMA-2 model with 70b parameters. Error bars represent 95\% CI across the 24 event pairs. }
\label{fig:prob_identity_bars}
\end{figure*}

\section{Evaluating Coherence in LLMs}

The probabilistic identities and repeated judgments used to study the coherence of human judgments provide an opportunity to study coherence in LLMs. To this end, we evaluated four distinct LLMs with varying levels of sophistication and model transparency. These included OpenAI's GPT-4 and GPT-3.5-turbo-1106 \parencite{achiam2023gpt}, as well as Meta's LLaMA-2-70b and LLaMA-2-7b \parencite{touvron2023llama}.

\subsection{Methods}
Each of these LLMs is equipped with a system message feature, which facilitates the application of instructions and ensures the consistency of constraints across different scenarios. For our experiment, we used a uniform system message for all probability queries, directed to each of the LLMs. The message instructed: ``\textit{You will estimate probabilities of some real-world events. Respond only with a number corresponding to a probability between 0 and 1. Do not respond with any other text. We are interested in your subjective evaluation of the probability so just respond what you think.}'' 

We presented LLMs with 24 pairs of constituent events, denoted $A$ and $B$ (detailed in Table \ref{tab:events}). The LLMs were tasked with assessing the probability of various combinations of these events, specifically $P(A)$, $P(B)$, $P(A\cap B)$, $P(A\cap\neg B)$, $P(B\cap \neg A)$, $P(A\cup B)$.
Of the 24 event pairs, 12 were related to weather conditions and the remaining 12 pertained to political scenarios. For the weather-related events, the prompts posed to the LLMs were framed as follows: ``\textit{What is the probability that the weather will be $X$ on a randomly-selected day in England during the year 2025?}''. In contrast, for the political events, the prompts were structured as: ``\textit{What is the probability that $X$ by the year 2025?}''. Appropriate logical operators, including negation (e.g., ``\textit{not $X$}''), conjunction (e.g., ``\textit{$X$ and $Y$}''), and disjunction (e.g., ``\textit{$X$ or $Y$}''), were applied to the pairs of events.

We configured the temperature parameter for all four LLMs to both 0 and 1. A zero temperature setting was selected to prompt each model to generate its most probable response. Conversely, a temperature setting of 1 was employed to make the LLMs to produce responses based on their learned probability distributions. To assess the variability of the responses, we submitted identical prompts to each LLM five times, with the temperature parameter maintained at 1 during these repetitions.

\subsection{Results}

\begin{figure*}[t!]
\centering
\includegraphics[width=\textwidth]{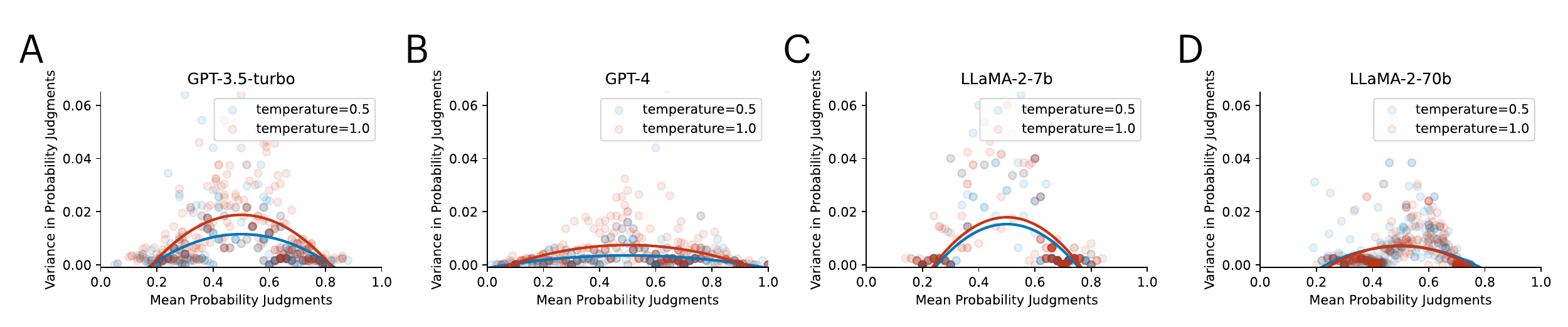}
\caption{The relationship between mean and variance in repeated probability judgments produced by LLMs exhibits an inverted-U shape. Solid lines represent the best-fitting regression models. Variants with more parameters or lower temperatures tend to shift the curve outward and downward, suggesting more consistent judgments. \textbf{(A)} GPT-3.5-turbo model. \textbf{(B)} GPT-4 model. \textbf{(C)} LLaMA-2 model with 7b parameters. \textbf{(D)} LLaMA-2 model with 70b parameters.  }
\label{fig:mean_var_relation}
\end{figure*}

As shown in Figure \ref{fig:prob_identity_bars}, the responses of LLMs exhibit characteristics akin to human reasoning biases. This similarity is primarily due to the way positive and negative terms within each identity collectively influence the extent of deviations from zero. The relationship between the balance of positive and negative terms and the magnitude of deviations from zero remains consistent across different types and variants of models. Moreover, the mean-variance relationships of repeated probability judgments produced by LLMs also resemble the human pattern (see Figure \ref{fig:mean_var_relation}). The consistency of reproducing human patterns across LLMs strongly implies that these statistical properties of LLM responses is an inherent characteristic of autoregressive models.



On average, model variants characterized by an increased number of model parameters exhibit reduced deviations from zero in probabilistic identities, a trend observed across both the GPT and LLaMA series. Additionally, variants with a greater number of model parameters display reduced variability and a less pronounced 'shrinkage effect,' as evidenced by the mean-variance curves shifting concurrently downward and outward (see Figure~\ref{fig:mean_var_relation}). These observed patterns suggest a correlation between increased model size and enhanced coherence, although it is important to note that perfect coherence is not achieved in any model.

\section{Theories of Human Probability Judgments}

Before we delve into possible accounts for the observed patterns in probability judgments produced by LLMs, it is instructive to revisit theories of human probability judgments due to the behavioral parallels observed between LLMs and people.

Human biases in probability judgments have been explained using two leading models: PT+N \parencite{costello2014surprisingly} and the Bayesian Sampler model \parencite{zhu2020bayesian}. Both models posit the existence of an underlying event probability, denoted $\theta$, which, while coherent, may not necessarily align with actual probabilities. For example, when judging $P(\text{head})$ and $P(\text{tail})$ for a fair coin, $\theta$ can manifest as both coherent and accurate, exemplified by $\theta_\text{head}=\theta_\text{tail}=0.5$. Alternatively, $\theta$ can be coherent yet inaccurate, such as $\theta_\text{head}=0.6, \theta_\text{tail}=0.4$, where the underlying probabilities deviate from the actual probabilities of 0.5 in a fair coin. The key to coherence, in this example, lies in the condition that $\theta_\text{head}=1-\theta_\text{tail}$, ensuring the internal consistency of logically related events.

Both PT+N and Bayesian Sampler models incorporate the concept of an approximation process, which involves drawing samples from the underlying event probability $\theta$. 
A crucial distinction between the two theories lies in the handling of these samples: the PT+N model posits that these samples are subject to corruption by independent noise factors \parencite{costello2014surprisingly}. Therefore, the PT+N's interpretation of the observed patterns in probabilistic identities hinges on the presence of additive noise within individual probability judgments. This model suggests that an increase in the additive noise is likely whenever there is an imbalance between positive and negative terms.

In contrast, the Bayesian Sampler model postulates that these samples are integrated with a prior, $p(\theta)$, reflecting the prior knowledge of the underlying event probability \parencite{zhu2020bayesian}:
\begin{align}
    p(\theta | x) = \frac{p(x | \theta) p(\theta)}{p(x)}
    \label{eq:BS_posterior}
\end{align}
where $p(x|\theta)$ represents an approximation of the underlying event probability. Furthermore, the mean probability judgment as predicted by the Bayesian Sampler model corresponds to the mean of the posterior distribution in Equation \ref{eq:BS_posterior}:
\begin{align}
\mathbb{E}[\theta] & =\int \theta p(\theta|x) d\theta  = \frac{\int \theta p(x | \theta) p(\theta) d\theta}{\int p(x | \theta) p(\theta) d\theta}
\label{eq:BS_posterior2}
\end{align}

Thus, the interpretation offered by the Bayesian Sampler model is grounded in the repeated use of the prior, $p(\theta)$, in each instance of probability judgment. According to this model, the imbalance between positive and negative terms in an probabilistic identity serves to magnify the influence exerted by the prior.

To further illustrate the mechanisms of the Bayesian Sampler model, let's consider a symmetric Beta prior for $p(\theta)$, denoted as $\text{Beta}(\beta,\beta)$, and a binomial distribution for $p(x|\theta)$, denoted as $\text{Bin}(N,\theta)$. That is, the agent is assumed to draw $N$ i.i.d.-samples from the underlying event probability $\theta$. The posterior as predicted by the Bayesian Sampler model is in the form of $\text{Beta}(\beta+x, \beta+N-x)$ and $x \sim \text{Bin}(N,\theta)$, which has mean and variance:
\begin{align}
    \mathbb{E}[\theta] & = \frac{N}{N+2\beta}\theta + \frac{\beta}{N+2\beta} \\
    \mathbb{V}[\theta] & = \frac{N}{(N+2\beta)^2}\theta(1-\theta)
\end{align}

Considering $Z_3$ as an example, the predicted mean values for the identity can be articulated as follows:
\begin{align}
    \mathbb{E}[Z_3] & = \mathbb{E}[\theta_A] + \mathbb{E}[\theta_{B\cap\neg A}] -\mathbb{E}[\theta_{A\cup B}] \\
    \begin{split}
    & = \Big[ \frac{N}{N+2\beta}\theta_A+\frac{\beta}{N+2\beta} \Big] \\
    & \quad + \Big[ \frac{N}{N+2\beta}\theta_{B\cap\neg A}+\frac{\beta}{N+2\beta} \Big] \\
    & \quad - \Big[ \frac{N}{N+2\beta}\theta_{A\cup B}+\frac{\beta}{N+2\beta} \Big]\\
    \end{split}\\
    & = \frac{N}{N+2\beta} (\theta_A+\theta_{B\cap\neg A}-\theta_{A\cup B}) + \frac{\beta}{N+2\beta}  \\
    & = \frac{\beta}{N+2\beta}
\end{align}

Given that the underlying event probabilities $\theta$ are coherent, it follows that $\theta_A+\theta_{B\cap\neg A}-\theta_{A\cup B}=0$. Therefore, it is the number of terms $\frac{\beta}{N+2\beta}$ that prompts the identities to deviate from zero, which is related to the imbalance of positive and negative terms within an identity.

In scenarios where the agent does not draw any samples for estimating the underlying event probability (namely, $N=0$), and the prior distribution of this probability is symmetric, the expected deviation from zero is $0.5$ for identities $Z_{3:6}$ and $1$ for identities $Z_{7:8}$. These values can be considered as the maximal limits of incoherence within probabilistic identities. This analysis has two key implications: (1) An agent demonstrating incoherence levels behind these extremes is indicative of a more accurate approximation of the underlying event probabilities; (2) Conversely, any factor impeding the accurate approximation of the underlying event probability will result in identities exhibiting deviations that converge towards these bounds.

The Bayesian Sampler model also predicts the inverted-U-shaped relationship between the mean and variance of repeated probability judgments \parencite{sundh2023unified}. By rearranging the predicted mean and variance, we derive the following relationship for the aforementioned example:
\begin{align}
    \mathbb{V}[\theta] = \frac{1}{N} \mathbb{E}[\theta] \Big(1-\mathbb{E}[\theta] \Big) - \frac{\beta(N+\beta)}{N(N+2\beta)^2}
\end{align}

In this equation, the inverted-U shape is a result of the quadratic relationship between mean and variance. As a corollary, an increase in the sample size $N$ (i.e., obtaining a more accurate approximation of the underlying event probability) is predicted to shift the curve downward. Furthermore, the shrinkage effect that shifts the curve inward is driven by the incorporation of the $\text{Beta}(\beta,\beta)$ prior. An increase in the value of $\beta$ (i.e., a stronger prior) is expected to cause an inward shift of the mean-variance curve.


\section{Connecting LLMs to Bayesian Inference}
We now consider a possible explanation for the systematic deviations from rationality observed in LLMs. Our approach begins with a formalization of the autoregressive process employed by LLMs in the generation of probability judgments.

Let $w_{n+1}$ be the model-predicted probability judgment and $w_{1:n}$ the query (e.g., ``What is the probability that Greece has left the European Union by the year 2025?''). We note that the process for generating probability judgments in autoregressive LLMs can be understood as $p(w_{n+1} | w_{1:n})$. Under the exchangeablility assumption\footnote{While this assumption might not be rigorously applicable at the token level in LLMs, it can nonetheless function as a valuable proxy within the semantic space to explain LLMs' responses \parencite{xie2021explanation, zhang2023deep}.}, we can rewrite the autoregressive objective as an implicit Bayesian inference process using de Finetti's theorem \parencite{korshunova2018bruno, zhang2023deep}:
\begin{align}
    p(w_{n+1} | w_{1:n}) = \int p(w_{n+1} | \theta) p(\theta | w_{1:n}) d\theta 
    \label{eq:de_finetti}
\end{align}
where $\theta$ represents the latent generative process underlying the data or, in our study, the underlying probability of the event in the query. Equation \ref{eq:de_finetti} links the concept of learning to fit the autoregressive distribution (i.e., the left hand side) with conducting Bayesian inference on the latent generative process underlying the data (i.e., the right hand side) \parencite{zhang2023deep}. In our context, it is specifically the underlying event probabilities on which the Bayesian inference is executed. 


This decomposition of the autoregressive distribution resembles the Bayesian inference process applied to samples in the Bayesian Sampler model:
\begin{align}
    w_{n+1} & = \arg\max_{w_{n+1}} p(w_{n+1} | w_{1:n}) \\
    & = \arg\max_{w_{n+1}} \int p(w_{n+1} | \theta) p(\theta | w_{1:n}) d\theta \\
    & = \arg\max_{\theta} p(\theta|w_{1:n}) \\
    & \approx  \int
    \theta p(\theta | w_{1:n}) d\theta  = \frac{\int \theta p(w_{1:n} | \theta) p(\theta) d\theta}{\int p(w_{1:n} | \theta) p(\theta) d\theta}
    \label{eq:LLM_BS_link}
\end{align}
where we assume in Equation 12 that $p(w_{n+1}=\theta|\theta)=1$ (i.e., $p(w_{n+1} \neq \theta|\theta)=0$)\footnote{LLMs were assumed to demonstrate truthful reporting of latent event probabilities.} and in Equation 13 that the mode is approximated by the mean of the integral. By contrasting Equation \ref{eq:LLM_BS_link} with \ref{eq:BS_posterior2}, we postulate a possible link between the probability judgments produced by LLMs and the Bayesian Sampler model. In this framework, the prompt $w_{1:n}$ acts as the target for approximating the underlying event probability in LLMs.


\section{Discussion}

We have presented empirical evidence demonstrating the incoherence of probability judgments generated by LLMs. Furthermore, through the analysis of probabilistic identities and mean-variance relationship, we identified shared patterns of incoherence in LLMs and humans. These structures offer insights into the underlying mechanisms employed by LLMs in the formation of probability judgments. We conjecture that this process originates from the implementation of autoregression for the four LLMs. Specifically, we postulated a possible connection between the autoregressive training objective and the Bayesian Sampler model, which has been previously employed to account for similar patterns of incoherence observed in human judgments \parencite{zhu2020bayesian, zhu2023autocorrelated, sundh2023unified}.

Our results suggest a novel approach for enhancing the accuracy of probability outputs generated by AI systems. This improvement might be achieved not through calibration with true frequencies, but rather by adjusting the degree of incoherence in the output. The established relationship between coherence and accuracy in probability estimation, particularly in the context of agents with bounded rationality \parencite{zhu2022clarifying, leitgeb2010objective}, provides a theoretical foundation for this method. Applying this relationship to recalibrate incoherent judgments offers a promising avenue for future research. We see this approach as having the potential to help refine AI-based probability judgments, making the resulting models more reliable and effective in practical applications.

Our work also exemplifies the idea that Bayesian and neural network models are complementary approaches in understanding machine and human intelligence \parencite{griffiths2023bayes}. The Bayesian model serves as a valuable tool for defining the computational-level objectives that a neural network model aims to achieve. This synergy between Bayesian and neural network models provides a more holistic understanding of intelligence, offering insights into the interplay between computational-level objectives and algorithmic-level execution in artificial and natural cognitive processes.

\vspace{2mm}

\noindent {\bf Acknowledgments.} 
This work and related results were made possible with the support of the NOMIS Foundation, as well as Microsoft Azure credits supplied to Princeton and a Microsoft Foundation Models grant. We thank Adam Sanborn for helpful discussion.

\printbibliography

\end{document}